# An Acceleration Method Based on Deep Learning and Multilinear Feature Space

Michel Andre L .Vinagreiro[1], Edson C. Kitani[2], Armando Antonio M. Lagana[1] and Leopoldo R. Yoshioka[1]

[1]Laboratory of Integrated Systems, Escola Politecnica da Universidade de Sao Paulo, Sao Paulo, Sao Paulo, Brasil
michel.vinagreiro@usp.br

[2]Department of Automotive Electronics, Fatec Santo Andre, Santo Andre, Sao Paulo, Brasil
edson.kitani@fatec.sp.gov.br

[1]Laboratory of Integrated Systems, Escola Politecnica da Universidade de Sao Paulo, Sao Paulo, Sao Paulo, Brasil
lagana@lsi.usp.br

[1]Laboratory of Integrated Systems, Escola Politecnica da Universidade de Sao Paulo, Sao Paulo, Sao Paulo, Brasil
leopoldo.yoshioka@usp.br

**ABSTRACT**

*Computer vision plays a crucial role in Advanced Assistance Systems. Most computer vision systems are based on Deep Convolutional Neural Networks (deep CNN) architectures. However, the high computational resource to run a CNN algorithm is demanding. Therefore, the methods to speed up computation have become a relevant research issue. Even though several works on architecture reduction found in the literature have not yet been achieved satisfactory results for embedded real-time system applications. This paper presents an alternative approach based on the Multilinear Feature Space (MFS) method resorting to transfer learning from large CNN architectures. The proposed method uses CNNs to generate feature maps, although it does not work as complexity reduction approach. After the training process, the generated features maps are used to create vector feature space. We use this new vector space to make projections of any new sample to classify them. Our method, named AMFC, uses the transfer learning from pre-trained CNN to reduce the classification time of new sample image, with minimal accuracy loss. Our method uses the VGG-16 model as the base CNN architecture for experiments; however, the method works with any similar CNN model. Using the well-known Vehicle Image Database and the German Traffic Sign Recognition Benchmark, we compared the classification time of the original VGG-16 model with the AMFC method, and our method is, on average, 17 times faster. The fast classification time reduces the computational and memory demands in embedded applications requiring a large CNN architecture.*

**KEYWORDS**

*Convolutional Neural Networks, Deep CNN model Acceleration, Advanced Driver Assistance Systems*

## 1. INTRODUCTION

The use of computer vision in Advanced Driver-Assistance Systems (ADAS) for environment mapping with images turns possible the recognition of persons, lane road, animals, vehicles, and traffic signs in real-time. The first algorithms designed for computer vision were based on image processing techniques, such as color segmentation, the histogram of oriented gradients, and cross-correlations. Image processing techniques show good performance for time operation and have an easy implementation. The methods' drawbacks are loss of performance in different light conditions, severe precipitation, mist, and occlusions. In this way, the necessity of robust solutions for ADAS environments rises and, the application of neural networks and Convolutional Neural Networks (CNNs) turns a new research field.

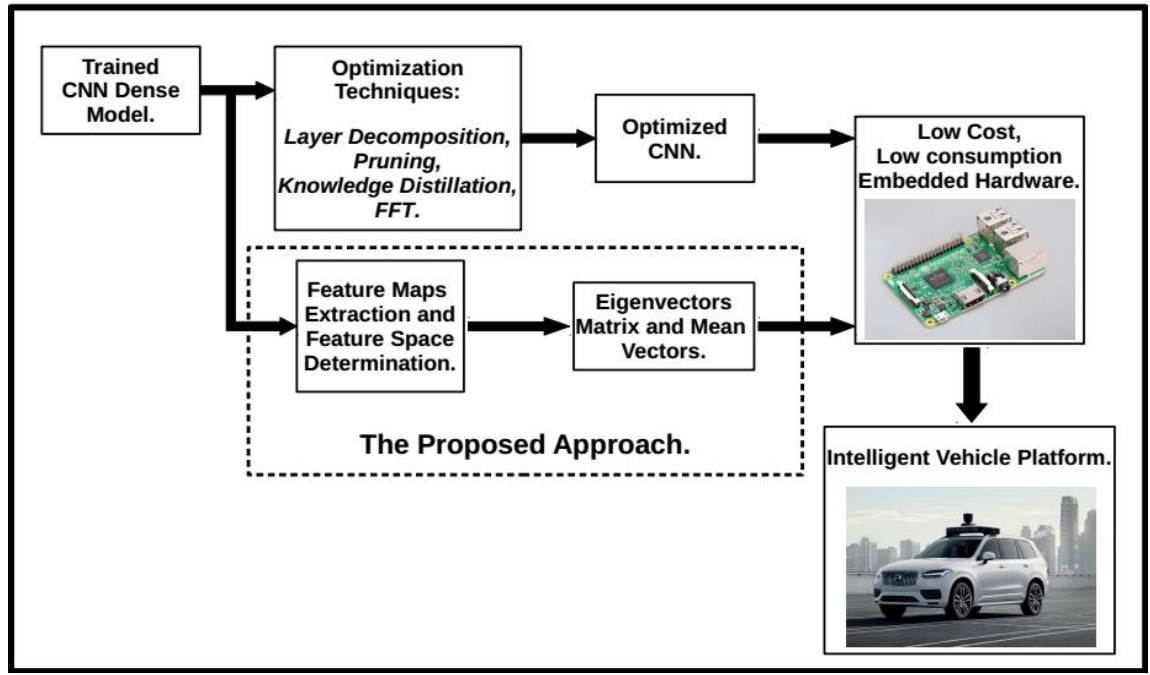

Figure 1. Illustration of the CNN reduction methods for ADAS applications. After train the dense model, the traditional methods reduce the complexity of the model. After the reduction, the embedded hardware platform host the minimal model. Our proposed framework does not reduce the model complexity. However, it uses the kernel knowledge present in the maps to determine the feature spaces and use it for the classification process.

At the end of the '90s, Lecun *et. al.* published in [1] and [2] the development and application of Convolutional Neural Networks (CNN). CNN is considered a Deep Learning algorithm and achieved the best performance in image recognition, localization, and segmentation tasks, compared with the traditional image processing techniques [3] and [4], mainly due to CNN's ability to extract a large number of features from input images. When Krizhevsky *et al.* [5] won the Imagenet-2012 challenge, the breakthrough occurred, significantly improving performance than previous architecture. Another successful architecture is the deep neural network proposed in [6], called VGG-16 (Visual Geometry Group), which showed the importance of depth architecture to achieve high-performance classification tasks.

Large-scale CNN networks such as VGG-16 are applicable in many classification tasks, including ADAS, mainly used for visual detection and mapping the environment. Computer vision is an essential subsystem composing ADAS systems, mainly used in vehicles for safety, lane-keeping, and collision avoidance systems. CNN is often used in self-driven cars to detect and recognize vehicles, persons, animals, and other obstacles. However, CNN's application for real-time operation requires more attention when running in vehicle's embedded platforms due to the need for high-spec hardware (RAM, CPU, and GPU). Some new approaches propose to deal with the real-time requirements as the problem mentioned above. One is developing CNN architectures with high performance and low computation cost [7] or compact and less powerful versions of large-scale architectures [8]. Other research lines focus on accelerating the classification time of large CNNs using strategies to optimize kernel activations [9]. The method uses the Single Value

Decomposition (SVD) as a low-rank approximation approach to accelerate the classification time of very deep CNNs. Other researches that present methods for acceleration of CNNs are [10] and [11]. In [12], the authors present a study on the relationship between operating speed and accuracy in CNN network applications used in object detection within an image. That work conducts a study of the balance between accuracy and time of operation through variations of characteristics of the architectures, such as extractors of features, resolution of the input images, etc. The study published in [13] proposes factorizing the convolutions into 2D kernels instead of 3D convolutions.

The work reports that the accuracy did not reduce severely and, the time of classification and training decreased a lot. The method proposed in [14] is an evolution of pruning methods for large CNN architectures [15]. The method's purpose is to use the PCA in the network analysis to discover and determine which kernels produce the largest variance results during the training process, thus reducing the accumulated error. Using those kernels and layers, the CNN model is retrained with a compressed version of the architecture. Figure 1 shows the applicability of CNN reductions methods and our proposed framework for the ADAS platform. Unlike the methods presented previously, this paper presents a new approach applied to any large-scale CNN architecture. It uses feature maps for determining the reduced dimensional space. Using this new space, we generate low-dimensional samples and train an external classifier. In figure 2 show our proposed method.

Despite the universality of our method, we will use the VGG-16 network as the basis for the experiments to validate our method's effectiveness. The rest of the paper is organized as follows: Section II describes a basic CNN structure and an overview of the PCA and MPCA methods applied to image pattern recognition; section III describes the proposed method. Section IV presents the experiments and discusses the results. Finally, section V presents the conclusion.

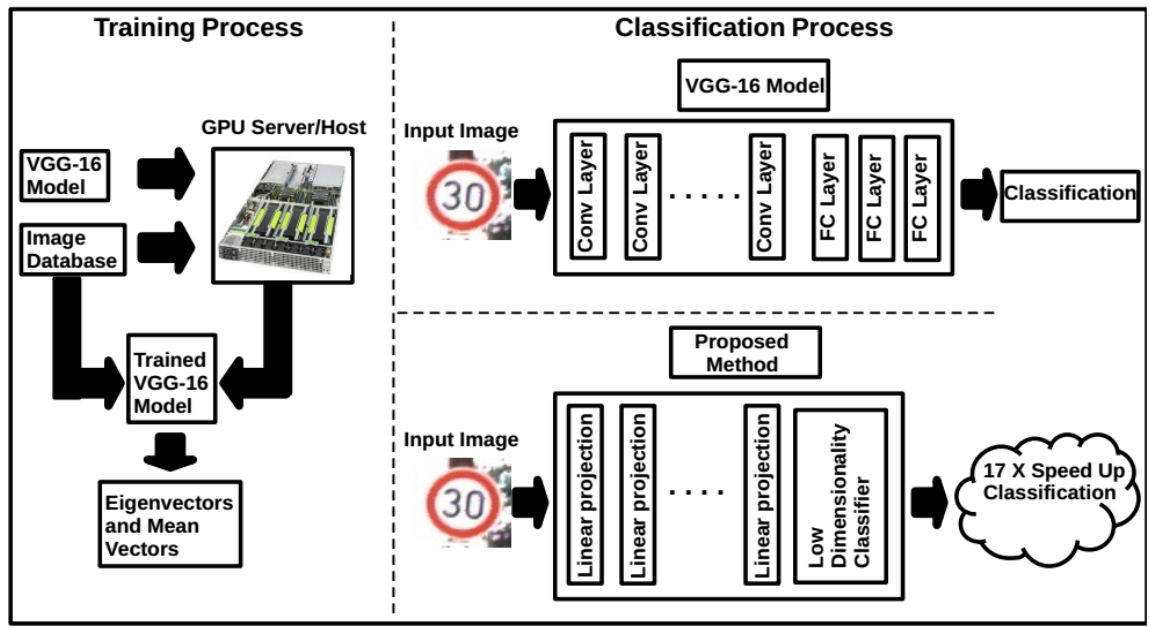

Figure 2. **Left side**: We train the VGG-16 model on the database with a cloud GPU server. With the trained model, we use the same database to extract banks of eigenvectors and mean vectors used to reduce the architecture. **Right side**: The VGG-16 model and the proposed method comparison: The substitution of the convolution processes by the chain of products accelerates the classification 17 times.

## 2. THEORY BACKGROUND

### 2.1. Convolutional Neural Networks

CNN structures are usually composed of three types of layers. The first, called the convolutional layer, can extract many features from images by convolution processes between regions of the input image and the layers' kernels. Every internal kernel element is an adjustable parameter adapted during the training phase, and the activation function determines the final output of the kernel [16]. The kernel slides by the whole image according to the parameters stride and padding. All the convolution process outputs are arranged in the feature map matrix [16], and the kernel for each convolutional layer generates the feature maps. The second layer, called sub pooling, uses the feature maps generated by previous convolutional layers. The regions of feature maps are sub-sampling, and the layer's output is a reduced dimension feature map. The operators of subsample can be the *maxpooling*, *meanpooling*, or *minpooling*. The maxpooling operator is the most used. Finally, the third layer, called fully connected (FC), consists of neuron units disposed of in interconnected multi-layers. The input of the first FC layer consist of all flattens feature maps from the last convolutional layer. The last FC layer can be a probabilistic function or a classifier, such as *Support Vector Machines* (SVM) or *Radial Basis Functions* (RBF).

### 2.2. PCA and MPCA

One of PCA's main applications [17] in image processing has been the dimensionality reduction of samples. Even though PCA has 120 years since Karl Pearson proposed it in 1901, it remains very current and helpful. Fundamentally, PCA creates a centered and orthogonal basis from the data's covariance matrix, maximizing the variance corresponding to the largest eigenvalues. This orthogonal basis maps the input data X into this new PCA space rotating the data distribution according to the highest eigenvector with to non-zero eigenvalue. Formally, PCA will find an orthogonal matrix $\boldsymbol{\Phi}$ that maps $\mathbb{X} \in \mathbb{R}^n$ to $\mathbb{Z} \in \mathbb{R}^p$, where p << n.

The eigenvectors of covariance matrix $\mathbb{X}$ are called *Principal Components* of set $\mathbb{X}$. The projection of any arbitrary $x$ sample into the new PCA feature space can be defined by $z = \boldsymbol{\Phi}^T x$, where $\boldsymbol{\Phi}$ is an orthogonal matrix whose the *kth* column is the *kth* eigenvector from the covariance matrix $\boldsymbol{\Sigma} = \boldsymbol{\Phi}\boldsymbol{\Lambda}\boldsymbol{\Phi}^T$ and $\boldsymbol{\Lambda}$ is the diagonal matrix whose $k$ is the *kth* eigenvalue of $\boldsymbol{\Sigma}$.

The idea behind the PCA is that the projection of any sample $x$ from the original space to the new PCA space will not change the original distribution once PCA is a linear approach based on the covariance matrix $\boldsymbol{\Sigma}$ of input matrix $X$. However, to deal with tensors in the CNN convolution layer, we need to consider a different approach, such as Multilinear PCA (MPCA) as proposed by [18].

Lu et al. [18] proposed Multilinear PCA (MPCA) for tensor objects as multidimensional objects, mainly related to videos and images. Considering a sequence of frames from a video file, $\mathbf{A} \in \mathbb{R}^{l1 \times l2 \times \ldots \times ln}$ will be the tensor object of nth-order and each frame $\mathbb{U}_{lk} \in \mathbb{R}^{i \times j}$, where k = 1, 2,…, N. Although, MPCA will reduce the total dimensionality from N ×i×j to P ×i×j, where P << N.

The MPCA requires a stack of input data $\mathbf{X}_k \in R^{i \times j}$ to project the tensor object $\mathbf{A}$ to the new reduced tensor space. The reduction occurs by the product of tensor $\mathbf{A}$ by a matrix $\mathbf{U} \in R^{in \times jn}$ denoted as $\mathbf{A} \times \mathbf{U}$, and $\mathbf{U}$ corresponds to the N projections matrices that maximize the M scatter of the tensors defined by $\psi_A = \Sigma_M \|\mathbf{A} - \mathbf{A_m}\|$, where $\mathbf{A_m}$ is the mean tensor.

### 2.3. Distilling the Knowledge in a Neural Network.

In [19], the authors propose a transfer learning method by training a minimal CNN architecture with the same training subset. The training labels of the minimal models are the soft activation outputs of the dense model. The computation of soft outputs uses the modified soft plus function using the temperature factor $T$, that's determinates the smooth of the outputs.

When the minimal model training starts, the value of T is high, decreasing with every iteration. In the end, the value of T is fixed in 1.

The objective error function of the minimal model is cross-entropy. When the temperature T is 1, the objective function is pondered by weight $T^2$. Thereby, the computed mean value defines the final output.

## 2.3. Pruning Convolutional.

The well-known pruning methods apply a unit dropout based on their importance on the final error of the network.

Let's $C(D/W)$ be accost function, $D=\{(x_1, x_2, ..., x_N),(y_1, y_2, ..., y_N)\}$ the training samples and labels and $W=\{(W^1,b^1), (W^2,b^2), ..., (W^L,b^L)\}$ is the weights set for all $L$ layers. Hence, for a set of weights $W^*$ resultant of pruning the function cost will be $C(D/W^*)$.

For each iteration, the less important unit is dropout. This process continues until reach the stopping criteria. The binary operator $g_l^k$ turns either off or on the unit $k$ of a layer $l$.

The work in [20] presents some pruning criteria. The *minimal norm* criteria compute the *l2-norm* for a set of weights for a unit. The unit with a low norm has a weak contribution for minimize the network error and can be dropped. The standard deviation criteria analysis the mean of the standard deviation of activation values for drop decision. The work cited the mutual information and Taylor expansion as alternative criteria.

## 3. THE PROPOSED METHODOLOGY

### 3.1. Definition

The proposed method adapted for VGG-16 is divided into five phases, as shown in the following:

• **Phase 1**: Initial step consists of applying the pre-processing to convert all images to the gray-scale and resize them to 224 × 224 pixels.

• **Phase 2**: The original VGG-16 model is trained with these pre-processed samples.

• **Phase 3**: $M$ image samples of the training subset, with $M < N$, are presented to the trained VGG-16 model and generates $K_l$ feature maps for each image in each layer $l = \{1, 2, 3,..., 13\}$ were $K_l$ is the number of kernels of the layer $l$. Each feature map is concatenated and arranged in the matrix $X^{(l)}$ of size $V \times n$, where $V$ is the product of $M$ per $K_l$ and $n = H^2$, where $H \times H$ is the input size. Before applying the PCA, the mean vector of $X^{(l)}$ is extracted and stored:

$$x_m^{(l)} = \frac{1}{V}\sum_{i=1}^{V} x_i \qquad (3.1)$$

The covariance matrix of $X^{(l)}$ is computed as:

$$Cov(x^{(l)}) = \frac{1}{V}\sum_{i=1}^{V}(x_i - x_m^{(l)})(x_i - x_m^{(l)})^T \qquad (3.2)$$

The $p_l$ eigenvectors of the X(l) covariance matrix related to non-zero eigenvalues compose the matrix $A^{(l)}$, with dimensionality $p_l \times n$. The matrix $A^{(l)}$ and the mean vector $x^{(l)}$ are phase three's output. Figure 3 illustrates all processes of phase 3.

• **Phase 4**: Step four consists of applying phase three for all layers of the model.

• **Phase 5**: In this phase, the downsize of training and test samples occur by their projections on layers space using the mean vectors and eigenvectors matrices. The training and validation processes of the low dimensional classifier use these low dimensional samples. Figure 4 illustrates phase 5.

For each layer $l$, the feature maps must be resized to $H \times H$, where $H = \sqrt{p_{(l-1)}}$, except for the first layer. This resize turns possible the process of the dot product that will generate the low dimensional samples.

In the dense models, the chain of subtractions and products using the matrices of eigenvectors and the mean vectors replaces convolutional processes. This replacement accelerates the time of classification.

The main objective of this work is to reduce the overall classification time for a new image sample. We call our proposed method as *Accelerated Multilinear Feature Space Classification technique* (AMFC).

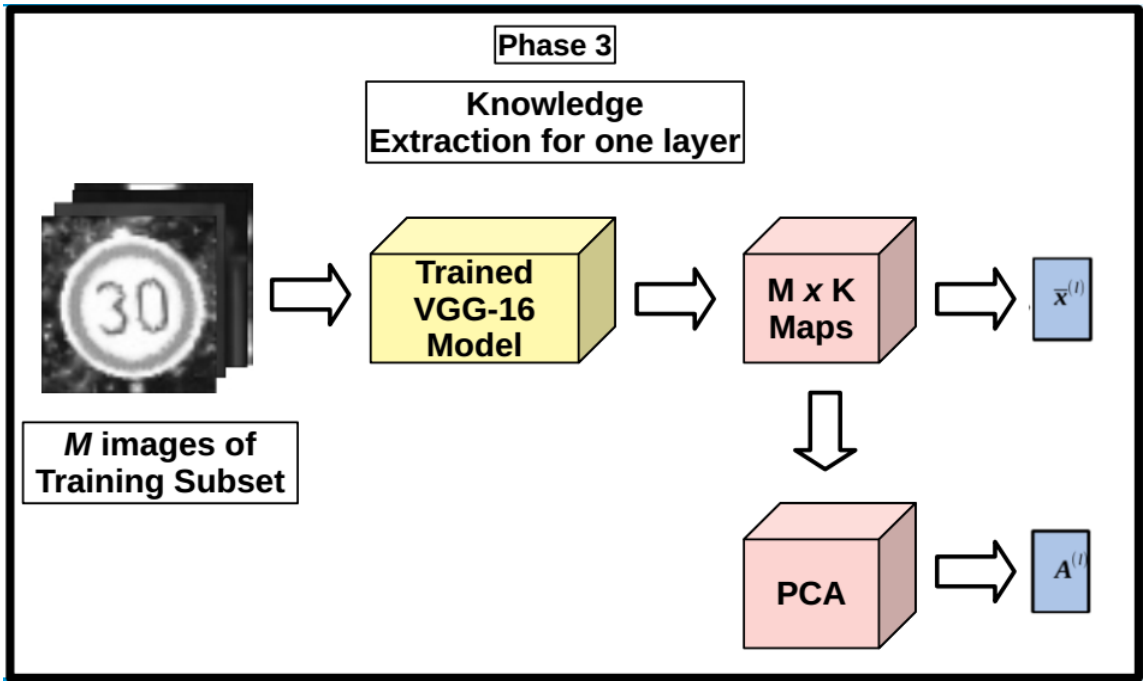

Figure 3. The illustration of the processes to obtain the matrix of eigenvectors and the mean vector for one layer $l$. In phase three, $M$ image samples are presented to the trained VGG-16

model to generate $M \times K_l$ features maps per layer. The mean vector $x^{(l)}$ and the matrix of eigenvectors $A^{(l)}$ are computed.

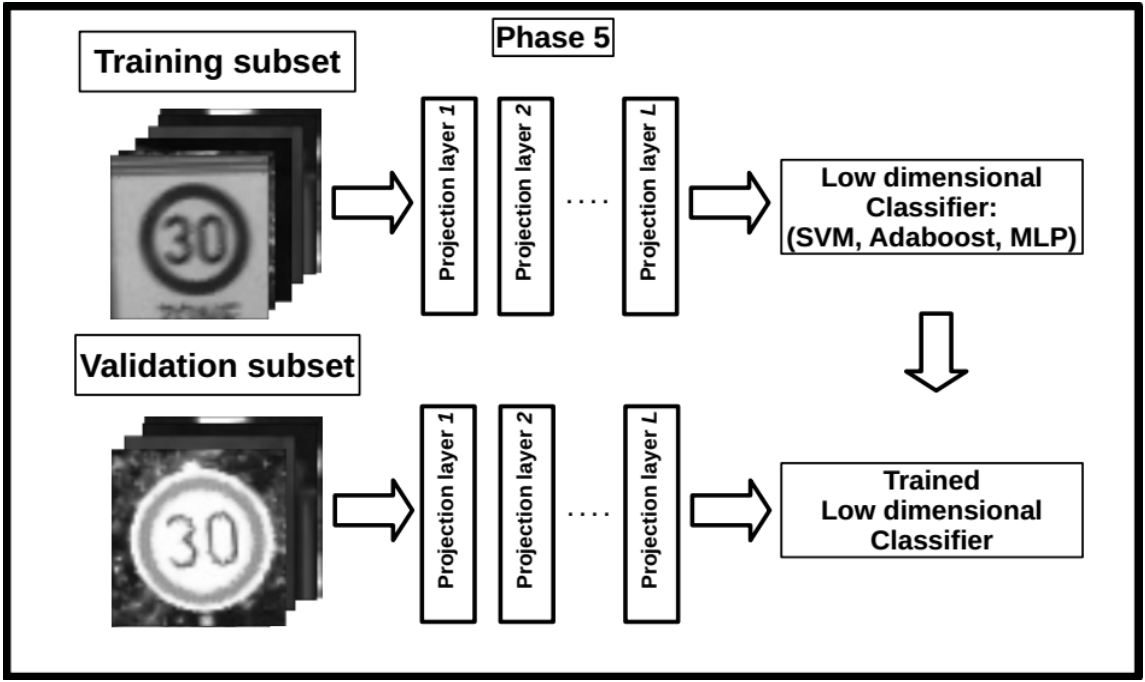

Figure 4. The illustration of the processes to project the samples on layers spaces. The projections generate the low dimensional samples that are used to train and validate the low dimensional classifier. After the end of the process, the classifier plays the fully connected layers function.

### 3.2. Samples with low dimensionality

Projecting any new image onto Feature Map Space requires resizing the image sample, *Imt*, in $224 \times 224$ pixels. In the second step, the new image is concatenated to vector *xt*, $1 \times n$, with $n = 50176$. The projection of *xt* into space of the first layer, $z(xt)^{(1)}$, $1 \times p_1$ occurs by the subtraction of mean vector $x^{(1)}$ and the dot product with $A^{(1)}$. The vector $z(xt)^{(1)}$ is projected into space of the second layer by the same process above, generating $z(xt)^{(2)}$, and then projected into space of the third layer, and repeating the process until the last layer as shown in equations 3.3, 3.4 and 3.5 respectively.

$$z(xt)^{(1)} = (xt - x_m^{(1)})(A^{(1)})^{(T)}$$
(3.3)

$$z(xt)^{(2)} = (z(xt)^{(1)} - x_m^{(2)})(A^{(2)})^{(T)}$$
(3.4)

$$z(xt)^{(L)} = (z(xt)^{(L-1)} - x_m^{(L)})(A^{(L)})^{(T)}$$
(3.5)

As mentioned early, the low dimensional samples are used to train and validate an external classifier that substitutes the fully connected layers of the VGG-16 model.

# 4. EXPERIMENTS AND RESULTS

A set of experiments were conducted to evaluate the capability of AMFC to speed up the classification time with minimal accuracy loss. The first experiments with six scenarios varying parameters were conducted. To exploit the best scenario, we used the cross-validation experiments at all scenarios and, the results were organized and presented in tables. To prevent overfitting, the training and validation of all classifiers use the *early stopping* method.

The difficulty of reproduction of CNN's reduction approaches turns impractical the use in experiments with these approaches. These implementations are crucial for comparing our proposed method and obtaining the proposed method's overall situation in the research area.

## 4.1. Databases

### 4.1.1. Vehicle Image Database

*Universidad Politécnica de Madrid* [21] to evaluate computer vision algorithms for automotive applications built it. This image database is composed of 7325 images of road lanes with the presence of vehicles or not. The images belonged to two classes and were collected under different angles, environments, and light conditions. The images have dimensions of 64 × 64 pixels. Because of the unbalance of the database, with a different number of images per class and corrupted images, we used only 5400 of 7325 available. The best result achieved in the training process of model VGG-16 was 98.7% accuracy in the test set after 31 epochs, a learning rate of $10^{-6}$, and a mini-batch size of 20.

### 4.1.2. German Traffic Sign Detection Benchmark (GTSDB)

This image database is available at *Institut Für Neuroinformatik* of *Ruhr-Universitat Bochum* [22]. The database contains more than 50,000 images of traffic signs distributed in 43 classes. Simultaneously, the images were captured in several environments, different angles of view, light conditions, and different dimensions. We randomly select four classes of images to conduct the experiments. The best result achieved in the training process of the VGG-16 model was 99.7% of accuracy in the test set at 24 epochs, with learning a rate of $10^{-6}$ and a mini-batch size of 100.

## 4.2. Experiments Scenario Description

The experiments consist of the training and test of an external classifier with the samples projected on layers. The speed-up (SPU) of the time for classification is measured by:

$$SPU = \frac{t_{VGG}}{t_{MFS\_CNN}} \qquad (4.1)$$

Where $t_{VGG}$ is the time of classification of an arbitrary sample by trained VGG-16 model and $t_{MFS\_CNN}$ is the time of classification by our proposed method.

Before initializing the experiments, we have to compute the low dimensional samples considering the first seven and, after this, all layers spaces. For each matrix $X^{(l)}$, a total number of eigenvectors is extracted. The total number of eigenvectors extracted for layer 1 was *V-1*, and, for remains, $p_{(l-1)}$. To compose the matrices of eigenvectors, it was select different numbers of eigenvectors $p_l$ for each layer. The first ranked eigenvectors chosen from each layer that produced the best result were: 6889, 6724, 4096, 3364, 2304, 2116, 1600, 1444, 1156, 1024, 900, 784, 676, from the first

to the last layer, respectively. We used different scenarios to conduct the experiments. The experiments use all or part layer spaces to obtain the final vector. Besides, different combinations of eigenvectors compose eigenvectors' matrix, using these vectors to train and validate the external classifier. In the following, we describe the scenarios used in the experiment.

The scenarios are summarized in Table I, and the results of each scenario are presented in tables IV to IX, respectively. Before starting the experiments to check if the method effectively speeds up the classification time, we conducted cross-validation experiments to define which classifiers achieve higher accuracy values.

Table 1. Experiments Scenario Description

| Scenario | Layers Selected | Eigenvectors Selection |
|---|---|---|
| 1 | All 13 layers | First Ranked |
| 2 | First seven layers | First Ranked |
| 3 | All 13 layers | Last Ranked |
| 4 | All 13 layers | Randomly |
| 5 | First seven layers | Last Ranked |
| 6 | First seven layers | Randomly |

### 4.2.1. Cross-Validation Experiments

Before the random selection for subsets mounting, we set the k parameter of the k-fold algorithm as five, which always reserves 20% of total samples to test.

For each k-fold round, the VGG-16 model is trained with *k -1* subsets designed for the training process and validate with remain. We used $M = 1000$ randomly selected samples from the training subsets for generating the feature maps. All samples of the training subsets generate the low dimensional samples to train the external classifier. The proposed method was validated with the low dimensional samples generated with the same image sample subset to validate the original VGG-16 model.

The experiments described in this section used the following external classifiers: *Adaboost, Decision Tree, K-Nearest Neighbour, Naive Bayes, Random Forest, Multi-Layer Perceptron, and SVM*.

The best cross-validation results were achieved considering scenario 1. Tables 2 and 3 presents the best results from each classifier using validation subsets.

In database 1 [21], the best value achieved by the Adaboost classifier occurred when the number of estimators was set to 200. In the KNN classifier, the best value for the *k* parameter for all folds was 1. The multilayer perceptron classifier has three layers. The first layer has 1024 units; the intermediary layer has 256 and, the output layer 2 units. The activation function for the hidden layers and the output layer are *Relu* and *softmax*, respectively. The learning rate was fixed in $10^{-4}$, and the mini-batch size was 20. The best accuracy value in the fold was achieved after 24 epochs. The SVM classifier utilized the *linear kernel*.

Table 2. Cross-Validation Accuracy Performance on Database 1, best results highlighted in bold.

| Classifier | Fold 1 | Fold 2 | Fold 3 | Fold 4 | Fold 5 |
|---|---|---|---|---|---|
| Adaboost | 93.0% | 92.2% | 91.9% | 92.9% | 92.1% |
| D. Tree | 85.6% | 83.1% | 84.6% | 84.4% | 84.4% |
| K-NN | 90.0% | 88.1% | 89.4% | 90.3% | 89.5% |
| MLP | 97.1% | 97.2% | 96.8% | 97.0% | **97.3%** |
| N. Bayes | 85.1% | 83.1% | 82.0% | 83.9% | 82.6% |
| R. Forest | 86.1% | 86.9% | 88.8% | 88.4% | 87.8% |
| SVM | 88.8% | 88.4% | 77.5% | 83.0% | 89.0% |
| VGG-16 | 97.8% | 98.4% | 97.5% | 97.6% | **98.8%** |

In the database 2 [22], due to a large amount of memory required to store eigenvectors' matrices, we randomly choose four classes of 43. The best value achieved by the Adaboost classifier occurred when the number of estimators was set as 200. In the KNN classifier, the best value for the *k* parameter for all folds is 1. The multi-layer perceptron classifier has three layers. The first layer has 1024 units. The intermediary layer has 1024 and, the output layer has four units. We used as activation function for the hidden layers and the output layer are *Relu* and *softmax*, respectively. The learning rate has fixed in $10^{-5}$, and the mini-batch size is 25. The best accuracy value in fold 1 occurred at 26 epochs. The SVM classifier utilized the *Radial Basis Function kernel*.

Table 3. Cross-Validation Accuracy Performance on Database 2, best results highlighted in bold.

| Classifier | Fold 1 | Fold 2 | Fold 3 | Fold 4 | Fold 5 |
|---|---|---|---|---|---|
| Adaboost | 93.6% | 93.7% | 93.2% | 92.2% | 91.8% |
| D. Tree | 87.6% | 88.3% | 89.1% | 87.5% | 88.5% |
| K-NN | 97.2% | 97.2% | 97.9% | 98.1% | 97.7% |
| MLP | **99.6%** | 99.2% | 99.2% | 99.5% | 99.2% |
| N. Bayes | 68.7% | 69.7% | 69.4% | 67.0% | 68.4% |
| R. Forest | 83.1% | 83.2% | 84.3% | 84.0% | 83.7% |
| SVM | 97.5% | 98.0% | 97.7% | 98.2% | 97.8% |
| VGG-16 | **99.7%** | 98.5% | 98.7% | 99.4% | 99.2% |

Comparing the results presented in the tables, the classifiers that achieved the best overall results were the MLP and SVM, except for the first image database, which Adaboost overcome SVM.

### 4.2.2. Speed-up Experiments

The accuracy values achieved in scenario 1 using all databases are closest to the original VGG model. Table 4 summarizes the results achieved in image databases 1 and 2.

Table 4. Speed-up Performance on Database 1 and 2 for Scenario 1, best results highlighted in bold.

| Database 1 | | |
|---|---|---|
| **Classifier** | **Accuracy** | **SPU** |
| AMFC-MLP | **97.3%** | **16.9** |
| AMFC-Adaboost | 93.0% | 17.1 |
| VGG-16 | **98.8%** | - |
| Database 2 | | |
| **Classifier** | **Accuracy** | **SPU** |
| AMFC-MLP | **99.6%** | **16.8** |
| AMFC-SVM | 98.2% | 16.8 |
| VGG-16 | **99.7%** | - |

As expected in scenario 1, the loss compared with VGG-16 is minimal. The minimal loss probably occurs by the use of ordered high representation eigenvectors. That produces high information integrity as related by various works that use the PCA method. However, the best performance of the Adaboost classifier overcome SVM will investigate further.

We can easily conclude that the acceleration compared with scenario number 1 is due to the reduced number of layers. The global augment of loss can suggest that the performance is related to the totality of layers used in the classification task. Despite scenario 3 use all layers, the selection of eigenvectors with less associated eigenvalues decreases the global performance. Using a random selection of eigenvectors in scenario 4 reduces performance smoothly, but both the accuracy and acceleration remain close to scenario 1. This minimal loss and high acceleration can indicate a high redundancy of eigenvectors.

We can observe that the selection of eigenvectors is irrelevant when the method uses only the first layers. However, we can conclude that the complete solution for architecture reduction uses all layer spaces. Although, understand the operation in the first layers may elevate the acceleration without increase the loss.

Table 5. Speed-up Performance on Database 1 and 2 for Scenario 2, best results highlighted in bold.

| Database 1 | | |
|---|---|---|
| **Classifier** | **Accuracy** | **SPU** |
| AMFC-MLP | 95.1% | 17.3 |
| AMFC-SVM | 83.6% | 17.5 |
| VGG-16 | **98.8%** | - |
| Database 2 | | |
| **Classifier** | **Accuracy** | **SPU** |
| AMFC-MLP | 96.2% | 17.2 |
| AMFC-SVM | 98.2% | 16.8 |
| VGG-16 | **99.7%** | - |

Table 6. Speed-up Performance on Database 1 and 2 for Scenario 3, best results highlighted in bold.

| Database 1 | | |
|---|---|---|
| **Classifier** | **Accuracy** | **SPU** |
| AMFC-MLP | 96.4% | 16.9 |
| AMFC-SVM | 86.6% | 16.5 |
| VGG-16 | **98.8%** | - |
| **Database 2** | | |
| **Classifier** | **Accuracy** | **SPU** |
| AMFC-MLP | 98.4% | 16.7 |
| AMFC-SVM | 96.6% | 16.6 |
| VGG-16 | **99.7%** | - |

Table 7. Speed-up Performance on Database 1 and 2 for Scenario 4, best results highlighted in bold.

| Database 1 | | |
|---|---|---|
| **Classifier** | **Accuracy** | **SPU** |
| AMFC-MLP | 96.9% | 16.9 |
| AMFC-SVM | 88.8% | 16.5 |
| VGG-16 | **98.8%** | - |
| **Database 2** | | |
| **Classifier** | **Accuracy** | **SPU** |
| AMFC-MLP | 98.7% | 16.7 |
| AMFC-SVM | 97.7% | 16.8 |
| VGG-16 | **99.7%** | - |

Table 8. Speed-up Performance on Database 1 and 2 for Scenario 5, best results highlighted in bold.

| Database 1 | | |
|---|---|---|
| **Classifier** | **Accuracy** | **SPU** |
| AMFC-MLP | 95.0% | 17.2 |
| AMFC-SVM | 83.6% | 17.3 |
| VGG-16 | **98.8%** | - |
| **Database 2** | | |
| **Classifier** | **Accuracy** | **SPU** |
| AMFC-MLP | 95.8% | 17.1 |
| AMFC-SVM | 95.8% | 17.4 |
| VGG-16 | **99.7%** | - |

Table 9. Speed-up Performance on Database 1 and 2 for Scenario 6, best results highlighted in bold.

| Database 1 | | |
|---|---|---|
| **Classifier** | **Accuracy** | **SPU** |
| AMFC-MLP | 95.2% | 17.1 |
| AMFC-SVM | 83.2% | 17.0 |
| VGG-16 | **98.8%** | - |
| **Database 2** | | |
| **Classifier** | **Accuracy** | **SPU** |
| AMFC-MLP | 96.1% | 17.0 |
| AMFC-SVM | 95.5% | 17.3 |
| VGG-16 | **99.7%** | - |

### 4.2.3. Eigenvectors Contribution.

To investigate the contribution of each eigenvector for the composition of the result, we adopt the representation concept [24].

A valid associated eigenvalue indicates the representation of an eigenvector. A valid eigenvalue must be real, non-zero, and positive. Hence, the sum of all eigenvalues that satisfies these conditions is one. The simple relation between a valid eigenvalue and the sum of all defines its information representation.

From the PCA technique, the organization of the eigenvectors occurs with respect to the order of the largest to smallest associated eigenvalue representation. Base on this, the curve for a set of eigenvectors of a layer shows the representation spreading of associated eigenvalues. For example, we plot a representation curve for a set of eigenvectors of layer 1 and using database 2, presented in figure 5.

From the figure, we observe the high spreading on the interval of the first 100 eigenvectors and, for the rest, the information representation decreases smooth.

Figure 6 presents the representation curve for the eigenvectors of layer 13 using database 2. In this case, the spreading occurs along to the first 200 eigenvectors and, in general, smoothly compared to layer 1 case.

For layer 1, the information representation spreads along the large number of eigenvectors extracted. The first 10,77% of valid eigenvetors means 98,88% of the representation. Therefore, the rest of eigenvectors have weak participation. For layer 13, 39,60% summed 98,71% of the representation. On the last layers, we observed a large representation spreading.

We converted to the image, the first 20 and the last 20 eigenvectors to visualize and understand. Figures 7 and 8 present the visual transformations for layer 1 and layer 13, respectively. In layer 1, the first eigenvectors sound like the original maps. The last ones appear to be just a mean noise. For layer 13, the first eigenvectors present abstracts representations, and the last ones appear noisily. However, presenting intrinsic patterns.

The difference between the two layers suggests that the eigenvectors that represent a defined pattern have a high information representation value. In this way, noisily eigenvectors present a low value. The largest spreading of layer 13 can explain the intrinsic patterns of their last eigenvectors. We note that the spreading increases layer-by-layer. However, a well comprehensive of the eigenvectors of the layers will be studied in future works.

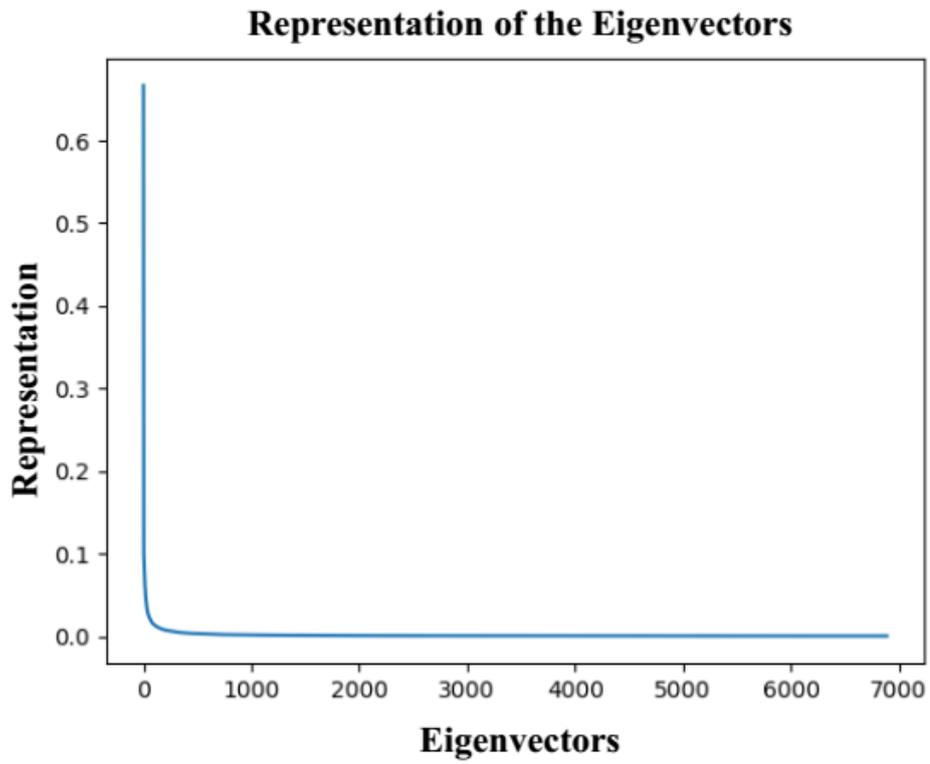

Figure 5. The representation curve for the first 7000 eigenvectors of layer 1 using database 1.

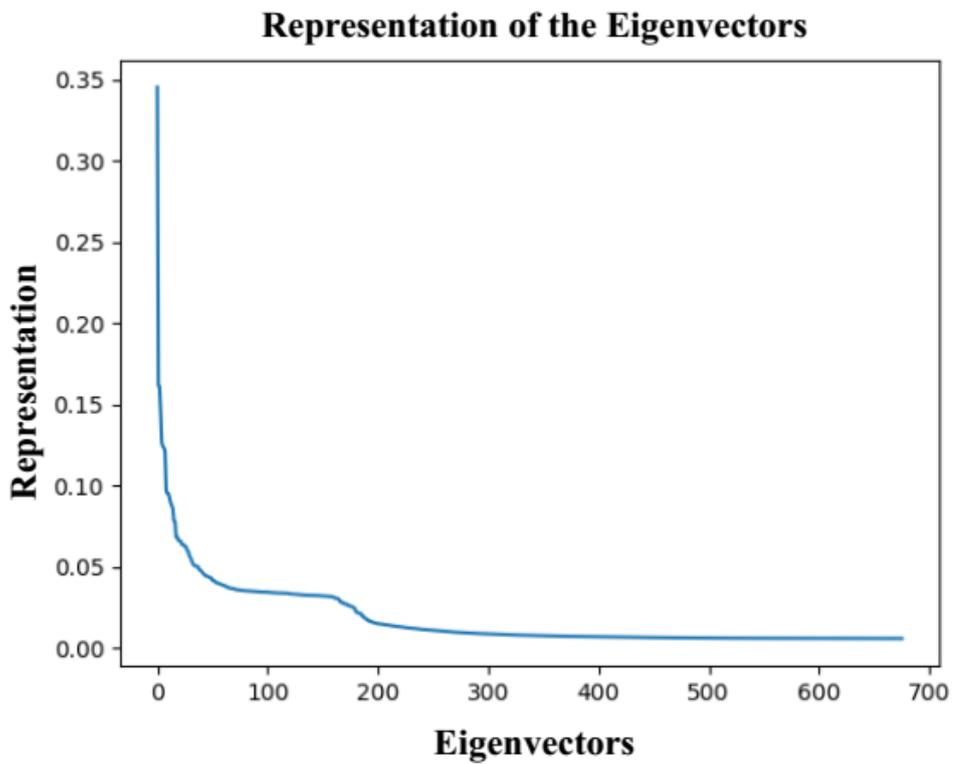

Figure 6. The representation curve for all eigenvectors of layer 13 using database 1.

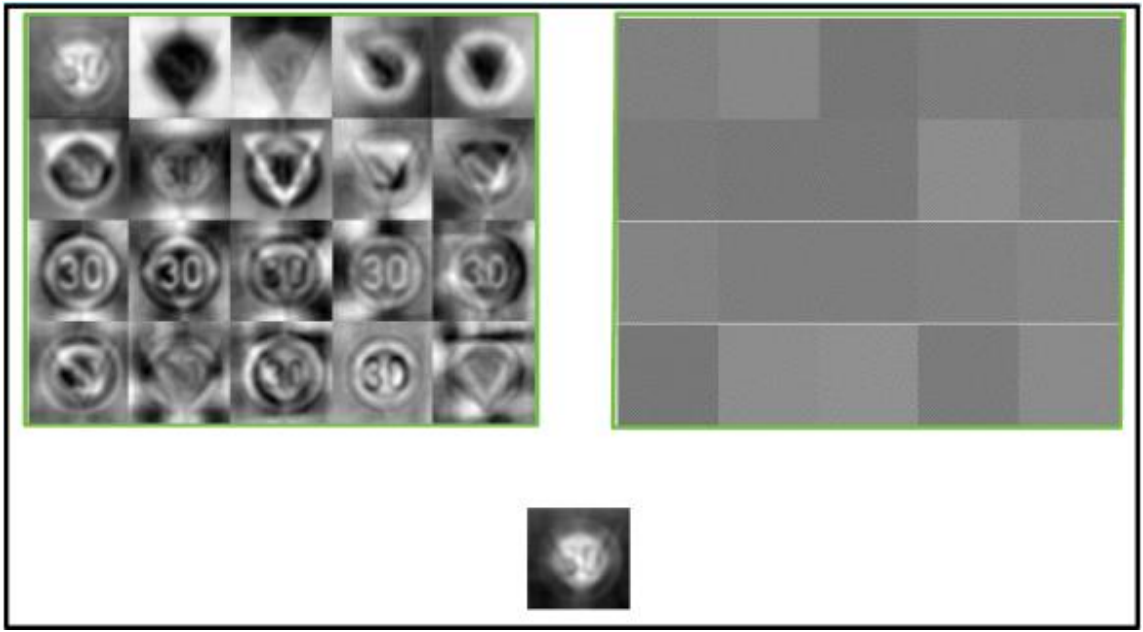

Figure 7. **Left side**: The reconstructed first 20 eigenvectors. **Right side**: The last 20. **Bottom**: The reconstructed mean vector of layer 1.

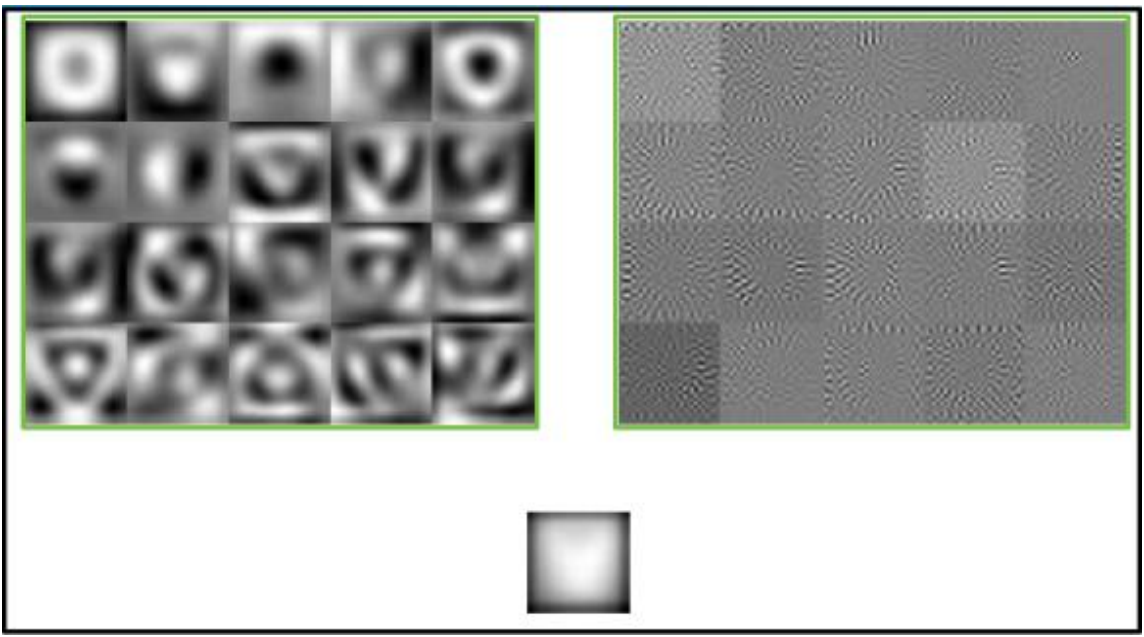

Figure 8. **Left side**: The reconstructed first 20 eigenvectors. **Right side**: The last 20. **Bottom**: The reconstructed mean vector of layer 13.

## 5. CONSIDERATIONS

The VGG-16 model with 10 classes uses approximately 1.6 GB of RAM. The expansion of memory occupancy occurs during the training process. When the database has many images, the training process on computers with limited memory space without GPU turns the process impractical.

The memory occupied by the proposed method occurs mainly due to the tensor of maps stored in the memory, with $H \times H \times M \times K_l$ bytes per layer, where $H$ represents the dimensions of maps. We observed that the growth of memory occupation is dependent on the number of $M$ samples. The value of $M$ needs to be great when the database has a large number of samples and classes. This rise is due to the necessity of representation of the total diversity of the database. Due to this drawback, the extraction of matrices of eigenvectors and mean vectors is infeasible when the image database has a large number of samples.

When the classification process of a new sample occurs in the VGG-16 model, the occupation of memory is due mainly to the storage of part of kernels weighs and the creation of the $Kl$ feature maps in the current layer in forwarding propagation mode. In the classification task, the proposed method occupies memories mainly with matrices of eigenvectors and mean vectors. The size of low dimensional samples is only of few kilobytes.

To perform the experiments, we used the *Google Colab service*. The service offers a cloud computing server with 32 GB of RAM and an *Nvidia Tesla K80 GPU*, *Nvidia Tesla T10,* or similar. The service was used only to train the original VGG-16 model. To extract the feature maps, compute the eigenvectors, train the external classifiers, and execute the test experiments, a personal computer with 8 GB of RAM and an Intel Core i5 Vpro processor was used.

All processes for extracting and storage the matrices of eigenvectors and mean vectors lasted six hours. The size of archives totaled 685 MB of RAM for ten classes. The proposed method achieved satisfactory results in the experiments but was not feasible with many classes and samples. This drawback is due to the high occupancy memory by the tensors.

Additionally, the method is ineffective when the objects of interest in the images have a high variance of size and position and are not aligned since the method is based on linear PCA.

Different works achieved good results in recent years by pruning [25] or compressing [26] large CNN architectures. However, our approach uses the ranked eigenvectors to reduce the classification time and not reduce the architecture's size.

## 6. CONCLUSIONS

In this paper, we presented an alternative method that focuses on the knowledges of CNN's kernels associated with a low complexity classifier to reduce the time of classification while preserving part of the performance reached by CNN.

The results have shown that AMFC is efficient in ADAS classification problems with a limited number of classes. The method is helpful in classification applications that use CNNs for embedded applications, with low computational resources in computer vision applications for the autonomous vehicle. The experiments with scenario 4 and 6 showed a reasonable accuracy with a high speed-up rate. In scenarios 4 and 6, we randomized the eigenvectors selection, even though the loss in accuracy was minimal. It is an indication that we have a high redundancy spread along all eigenvectors.

In the next step of this research, we will extend the application for other ADAS problems, such as license plate and vehicle type classification. The low consumption of the method turns the implementation and operation appropriate to the vehicular low-cost embedded platforms. These platforms are used mainly for performing real-time computer vision tasks.

In addition, we will evaluate a method to choose the minimum amount of the most significant eigenvectors, not considering only the eigenvalues as mentioned in this work, but the accuracy and reduced time for classification. The new version of AMFC will handle reasonably many samples and classes, outperforming the current drawback.

**Michel André L. Vinagreiro** is graduate in engineering, master's student and researcher with theLaboratory of Integrated Systems, Escola Politecnica da Universidade de Sao Paulo, Sao Paulo, Sao Paulo, Brasil.

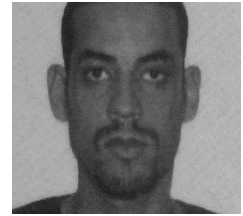

**Edson C. Kitani** has P.h.D degree by Escola Politecnica da Universidade de Sao Paulo and graduation professor in Fatec Santo Andre, Santo Andre, São Paulo, Brasil.

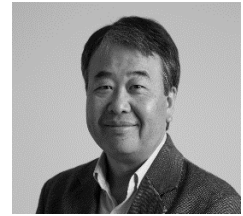

**Armando Antonio M. Lagana** has P.h.D degree by Escola Politecnica da Universidade de Sao Paulo, graduation professor and researcher with the Laboratory of Integrated Systems, Escola Politecnica da Universidade de Sao Paulo, Sao Paulo, Brasil.

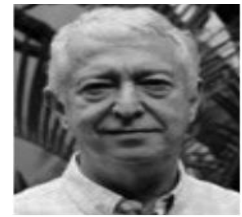

**Leopoldo R. Yoshioka** has P.h.D degree by Tokyo Institute of Technology, graduation professor and researcher with the Laboratory of Integrated Systems, Escola Politecnica da Universidade de Sao Paulo, Sao Paulo, Sao Paulo, Brasil.

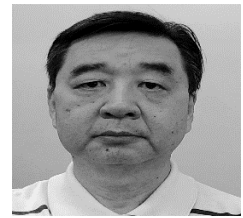